\documentclass[final]{cvpr}

\usepackage{times}
\usepackage{epsfig}
\usepackage{graphicx}
\usepackage{amsmath}
\usepackage{amssymb}
\usepackage{booktabs}
\usepackage{subcaption}
\usepackage{bbm}
\usepackage{mathrsfs}
\usepackage{soul}

\usepackage[table]{xcolor} % loads also »colortbl«

\usepackage{float}
\usepackage{tabulary,multirow,overpic,xcolor}
\usepackage{IEEEtrantools}

\newcommand{\Ours}{ILA-DA}

\newcommand{\bd}[1]{\textbf{#1}}

\usepackage{algorithmicx}
\usepackage{algorithm}
\usepackage[noend]{algpseudocode}
\usepackage{mathtools} 

\usepackage{enumitem}

\newenvironment{tight_itemize}{
\begin{itemize}[leftmargin=15pt]
  \setlength{\topsep}{0pt}
  \setlength{\itemsep}{2pt}
  \setlength{\parskip}{0pt}
  \setlength{\parsep}{0pt}
}{\end{itemize}}

\newcommand{\G}{\mathcal{G}}
\newcommand{\C}{\mathcal{C}}
\newcommand{\D}{\mathcal{D}}
\newcommand{\LL}{\mathcal{L}}
\newcommand{\Aff}{\mathrm{A}}
\usepackage{graphicx}
\usepackage[export]{adjustbox}

\usepackage{blindtext}
\usepackage[pagebackref=true,breaklinks=true,colorlinks,bookmarks=false]{hyperref}

\def\cvprPaperID{} % *** Enter the CVPR Paper ID here
\def\confYear{CVPR 2021}

\begin{document}
% \newfloatcommand{capbtabbox}{table}[][\FBwidth]

\definecolor{gray45}{rgb}{0.2, 0.5, 0.478}
%%%%%%%%% TITLE
\title{Instance Level Affinity-Based Transfer for Unsupervised Domain Adaptation}

\author{

\begin{tabular}{ccc}
 Astuti Sharma\hspace{20 mm} & Tarun Kalluri\hspace{20 mm} & Manmohan Chandraker
\end{tabular}

% \IEEEauthorblockN{} \\
 \\[2mm]
University of California San Diego \\
{\tt\small \{asharma,sskallur,mkchandraker\}@eng.ucsd.edu}

}
\maketitle
%%%%%%%%% BODY TEXT
\pagenumbering{gobble}
\begin{abstract}
Domain adaptation deals with training models using large scale labeled data from a specific source domain and then adapting the knowledge to certain target domains that have few or no labels. Many prior works learn domain agnostic feature representations for this purpose using a global distribution alignment objective which does not take into account the finer class specific structure in the source and target domains. We address this issue in our work and propose an instance affinity based criterion for source to target transfer during adaptation, called \Ours{}. We first propose a reliable and efficient method to extract similar and dissimilar samples across source and target, and utilize a multi-sample contrastive loss to drive the domain alignment process. \Ours{} simultaneously accounts for intra-class clustering as well as inter-class separation among the categories, resulting in less noisy classifier boundaries, improved transferability and increased accuracy. We verify the effectiveness of \Ours{} by observing consistent improvements in accuracy over popular domain adaptation approaches on a variety of benchmark datasets and provide insights into the proposed alignment approach. {Code will be made publicly available at \url{https://github.com/astuti/ILA-DA}.}
\end{abstract}

\section{Introduction}

\begin{figure}[t]
    \includegraphics[width=0.5\textwidth]{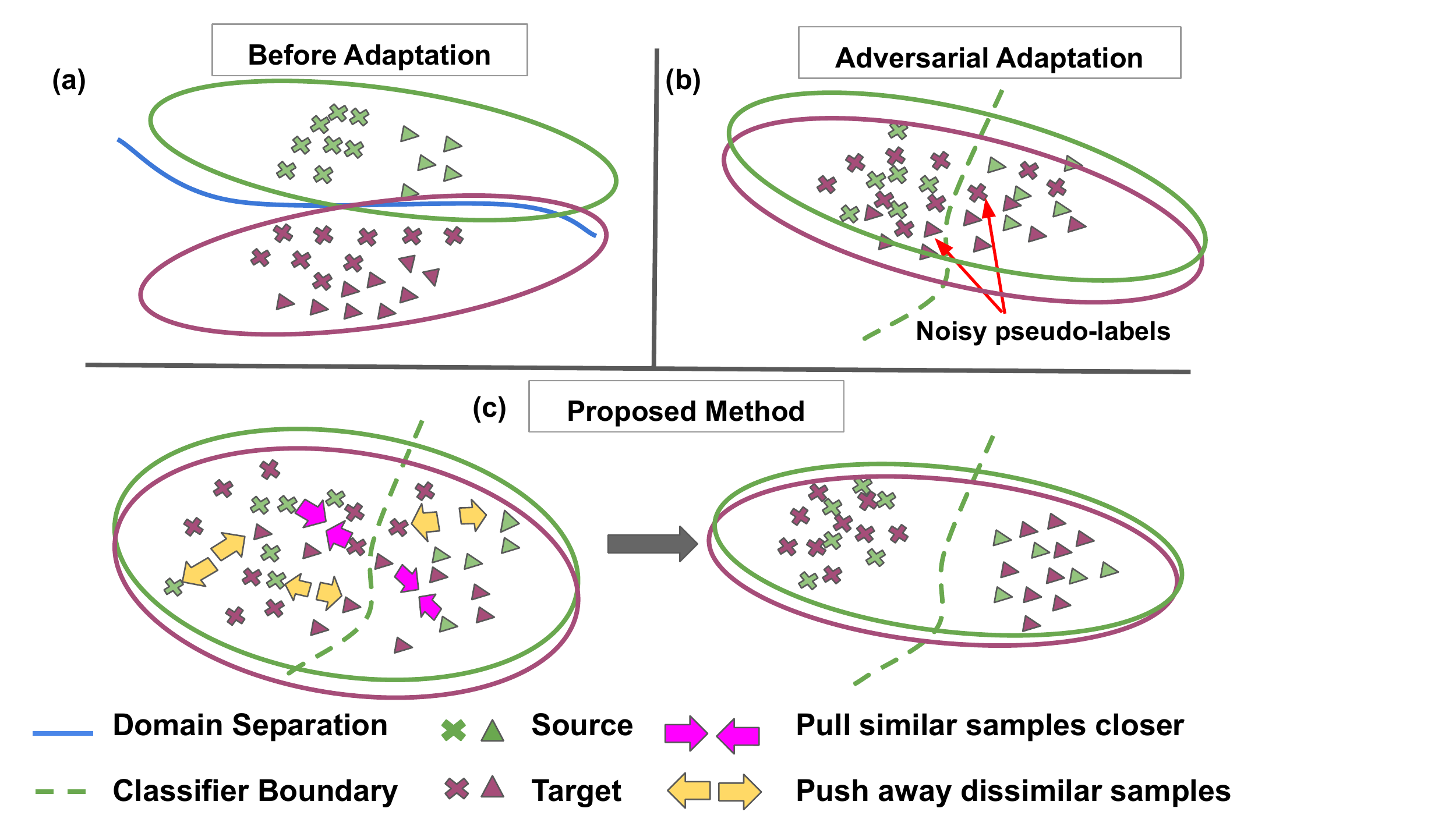}
  \caption{\textbf{Motivation for the proposed approach} (\textbf{a}), (\textbf{b}) Most adversarial learning based adaptation approaches achieve global domain alignment which often leads to misalignment near the classifier boundaries. (\textbf{c}) Using our affinity matrix based approach in combination with the proposed MSC loss, we achieve better discrimination between target samples and improve the adaptation.
}
  \label{fig:introPic}
\end{figure}

In this work, we propose a method to leverage instance wise similarities across datasets, called \Ours{}, to improve unsupervised domain adaptation.  

It is well known that models trained on a large-scale labeled dataset are generally sensitive to domain shifts and do not generalize well to data that lies outside the training distribution~\cite{torralba2011unbiased}. 
%%%%
Unsupervised domain adaptation~\cite{ben2006analysis, ben2010theory, saenko2010adapting} emerged as a feasible alternative to transfer knowledge from a labeled source domain to one or more unlabeled target domains by minimizing some notion of divergence between the domains~\cite{long2017deep,long2015learning,sun2016deep,DANN,tzeng2017adversarial, bousmalis2016domain}. A majority of successful approaches rely on global distribution alignment using adversarial learning~\cite{DANN, tzeng2017adversarial, bousmalis2016domain, bousmalis2017unsupervised, sener2016learning}, where the objective is to learn features that are good enough to fool a discriminator into classifying source samples as target and vice versa.
A major limitation with these methods is that while learning domain agnostic feature representations, they do not consider the finer class specific structure of the samples during the alignment resulting in noisy predictions near classifier boundaries. They do not take into account, for example, the fact that the affinity of different categories across source and target towards alignment can be different, which might lead to misalignment of few categories as shown in ~\figref{fig:introPic}. This problem is alleviated to an extent by many follow-up works that make use of target pseudo labels to guide class specific alignment \cite{kang2019contrastive, haeusser2017associative, long2013transfer, pei2018multi, luo2019taking, xie2018learning, pan2019transferrable}. However, the performance of these approaches is in most cases tied to the reliability of predicted pseudo labels which can be noisy without adequate filtering measures, leading to negative alignment between unrelated categories.

In this work, we address these limitations by proposing a novel adaptation approach called ILA-DA (Instance Level Affinity-based Domain Adaptation). We combine ideas from metric learning literature~\cite{weinberger2009distance, wang2018cosface, liu2017sphereface, deng2019arcface, deng2019arcface, qian2019softtriple, hadsell2006dimensionality} to perform cross domain transfer by using instance affinity relations between the source and target samples. As opposed to prior works that perform domain level or class level alignment, we show that a much finer knowledge in the form of sample level similarity can be successfully exploited to improve the adaptation process. The main challenge with this approach is that the target domain is completely unlabeled to extract similarity. To overcome this, we propose a nearest neighbor based technique to first construct a pairwise affinity matrix. We then use this knowledge of cross domain positive and negative relations in a multi-sample contrastive learning (MSC) loss that uses multiple positives and negatives across domains in a contrastive learning framework~\cite{hadsell2006dimensionality, oord2018representation}.

We identify two advantages using ILA-DA. Firstly, the pairwise similarities provide a relatively stronger signal during training and are shown to be more robust to label corruptions compared to category predictions in many cases~\cite{hsu2015neural, hsu2019multi}. Secondly, our multi-sample contrastive loss aims to cluster similar samples from across domain closer together while pushing dissimilar samples away to avoid negative transfer.
This is especially useful in adaptation across fine-grained datasets, where the challenge, apart from domain shift, is to additionally acknowledge the large intra-class variation within the categories. 

The effectiveness of ILA-DA is reflected by improved adaptation accuracy on popular benchmarks like Digits and Office-31 datasets. We also achieve state-of-the-art results on a challenging adaptation dataset Birds-31~\cite{PAN} without using complementary information such as label-hierarchies and class structure unlike \cite{PAN}, which indicates the usefulness of our MSC loss in handling wide variety of scenarios. We further perform extensive ablations and analysis on our methodological choices. All code and data for our method and baselines will be publicly released.

In summary, the key highlights of the paper are:
\vspace{-0.2cm}
\begin{tight_itemize}
    \item We propose a novel adaptation frame work ILA-DA. It uses Multi-Sample Contrastive (MSC) loss to perform instance affinity aware transfer by identifying pairwise similarity relations across source and target domains.
    \item ILA-DA is designed to be general and can be applied to enhance any existing adversarial adaptation approach. We show experimental results while using it in combination with two popular methods, DANN~\cite{DANN} and CDAN~\cite{CDAN}, and observe consistent improvements over both the baselines.
    \item We validate the effectiveness of the proposed approach numerically by applying it on multiple tasks from various challenging benchmark datasets used for domain adaptation like Digits, Office-31 and Birds-31 and observe improved accuracies in all the cases, sometimes outperforming the state-of-the-art by a large margin.
\end{tight_itemize}

\section{Related Work}
\label{sec:related_work}

\vspace{-0.2cm}
\paragraph{Domain Adaptation} 
Unsupervised domain adaptation enables training networks on completely unlabeled data by transferring knowledge from a model trained on a different labeled source domain. This is done by minimizing some notion of distance or divergence between the domains~\cite{ben2006analysis, ben2010theory}. The various notions of divergence include Maximum Mean Discrepancy (also known as MMD)~\cite{long2015learning, long2017deep, ming2015unsupervised, yan2017mind, baktashmotlagh2016distribution, pan2010domain, long2013transfer, tzeng2014deep, zhong2009cross, kang2019contrastive} between the feature embeddings of the source and target domains in a RKHS, higher order correlations between the domains~\cite{sun2015return, sun2016deep, morerio2017minimal}, optimal transport distance between the source and target~\cite{courty2016optimal, bhushan2018deepjdot} and distribution matching 
using generative~\cite{taigman2016unsupervised, sankaranarayanan2018generate, hoffman2018cycada} or discriminative~\cite{DANN, tzeng2017adversarial, bousmalis2016domain, bousmalis2017unsupervised, sener2016learning, CDAN} adversarial learning between a feature generator and a discriminator. In this paper, we propose complementary improvements to adversarial methods.

\vspace{-0.3cm}
\paragraph{Class-Specific Adaptation} 
Most of the above works aim to learn domain agnostic feature representations from the source and target data by aligning their global distributions, so that a source classifier can be used on the target. However this does not guarantee alignment between the respective categories which might lead to negative transfer.
Recent works alleviated this problem by taking into account class specific properties during adaptation between the domains \cite{bruzzone2009domain, saito2017asymmetric, long2013transfer, sener2016learning, zhang2018collaborative, luo2019taking, kang2019contrastive, xie2018learning, pan2019transferrable}. Since the target domain is completely unlabeled, these works rely on training co-regularization networks \cite{saito2017asymmetric, kumar2018co}, predicting psuedo-labels \cite{luo2019taking, kang2019contrastive, chen2019progressive} or computing prototypical \cite{snell2017prototypical} representations of source and target categories \cite{xie2018learning, pan2019transferrable, pinheiro2018unsupervised} to assign target classes during training. This makes the performance of these methods dependent on the pseudo-labeling hypothesis, leading to noisy predictions near the classifier boundaries. This is problematic, for example, in fine grained classification setting where the variation within a class is often large. In contrast, we propose a novel sample level transfer criterion which is robust to noisy psuedo-labeling and improves adaptation. A related work is Contrastive Adaptation Network \cite{kang2019contrastive}, but it is based on MMD and requires k-means clustering after each iteration to update pseudo labels, whereas our \Ours{} is an adversarial approach that uses the proposed affinity matrix combined with a new MSC loss to explicitly model pairwise interactions.

%%%%%%%%%%%%%%%%%%%%%%%%%%%%%%%%%%%%%%%%%%%%%%%%%%%%%%%%%%%%%%%%%%%%%%%%%%%%%%%%%%%%%%%%%%%%%%%%%%%%%%%%%%%%%%%%
%%%%%%%%%%%%%%%%%%%%%%%%%%%%%%%%%%%%%%%%%%%%%%%%%%%%%%%%%%%%%%%%%%%%%%%%%%%%%%%%%%%%%%%%%%%%%%%%%%%%%%%%%%%%%%%
\begin{figure*}[tbp]
  \centering
    \includegraphics[width=1\textwidth]{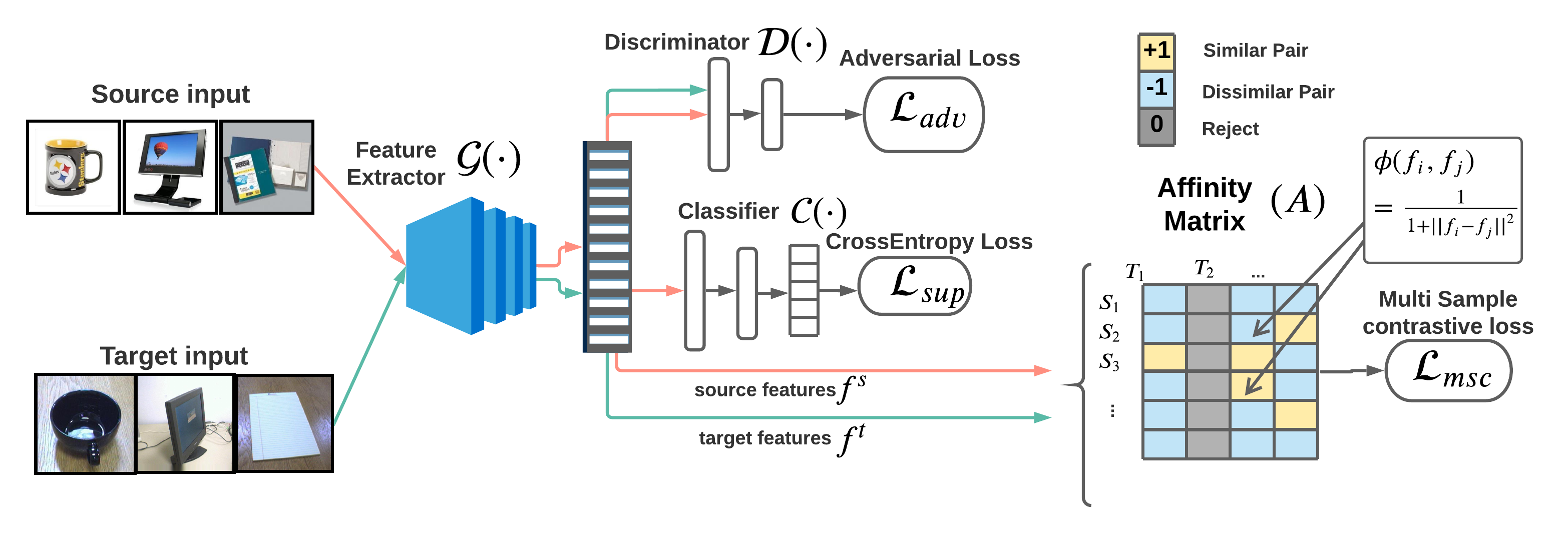}
  \caption{\bd{Illustration of proposed ILA-DA approach with MSC Loss.} Our architecture consists of a feature extractor $\G(.)$ that is shared across source and target domains. The classifier $\C(.)$ is trained to classify the source images using cross entropy loss $\LL_{sup}$, while the domain discriminator $\D(.)$ performs domain alignment using adversarial loss $\LL_{adv}$. Additionally, we use source and target features to construct an affinity matrix $\Aff$ that holds similarity and dissimilarity relations between the samples (\secref{subsec:affinity}). We then use this information to cluster categories closer to each other using our proposed multi-sample contrastive loss (\secref{subsec:msc_loss}). }
  \label{fig:full_model}
\end{figure*}
%%%%%%%%%%%%%%%%%%%%%%%%%%%%%%%%%%%%%%%%%%%%%%%%%%%%%%%%%%%%%%%%%%%%%%%%%%%%%%%%%%%%%%%%%%%%%%%%%%%%%%%%%%%%%%%
%%%%%%%%%%%%%%%%%%%%%%%%%%%%%%%%%%%%%%%%%%%%%%%%%%%%%%%%%%%%%%%%%%%%%%%%%%%%%%%%%%%%%%%%%%%%%%%%%%%%%%%%%%%%%%%

\vspace{-0.3cm}
\paragraph{Metric Learning} There have been a number of approaches proposed to learn discriminative boundaries between categories using sample-wise~\cite{weinberger2009distance, schroff2015facenet, oh2016deep, sohn2016improved, wang2019multi, kim2019deep, yu2019deep} or proxy-based~\cite{aziere2019ensemble, qian2019softtriple, movshovitz2017no, kim2020proxy} metric losses for tasks like face recognition~\cite{liu2017sphereface, wang2018cosface, deng2019arcface},
where the challenge is to concurrently address large intra-class variation as well as small inter-class differences. Our multi-sample contrastive (MSC) loss is built on top of the noise contrastive loss \cite{mnih2013learning} and softmax constrastive loss \cite{oord2018representation, henaff2019data}, where we extend it to handle multiple positives and negatives at once to 
leverage sample level relationships useful for adaptation. 

\vspace{-0.3cm}
\paragraph{Metric Learning for UDA} 
While there have been prior works that propose adaptation algorithms for metric learning \cite{sohn2018unsupervised, geng2011daml, ding2016robust}, there have been very few prior works that study the complementary problem of leveraging principles from metric learning to improve regular domain adaptation. Prior works either use triplet loss~\cite{laradji2020m} requiring complex sampling strategy or do not leverage instance level relations~\cite{pinheiro2018unsupervised}.In our work, we acknowledge the need to address intra-class variance within aligned source and target categories for adaptation, which we achieve by proposing a sample level cross dataset transfer mechanism.
\section{Proposed Method: ILA-DA}
\label{sec:method}

In this section, we first give a brief overview of adversarial adaptation methods, and then introduce our multi sample contrastive (MSC) loss for adaptation followed by construction of affinity matrix.

\subsection{Overview of Adversarial Domain Adaptation}

In the problem of unsupervised domain adaptation, we have a labeled source dataset $\mathcal{D}^s{:} \{ x_i^s , y_i \}_{i=1}^{|\mathcal{D}^s|}$, where $\mathcal{D}^s {\sim} P_s$ along with an unlabeled target domain $\mathcal{D}^t {:} \{ x_i^t \}_{i=1}^{|\mathcal{D}^t|}$ where $\mathcal{D}^t {\sim} P_t$, and $P_s \neq P_t$. The task is to train a model using these data to make predictions on $\mathcal{D}^t$. We present the overview of the architecture used for training in \figref{fig:full_model}. The feature extractor $\G$, which is shared between the source and the target images, extracts the lower dimensional feature representations corresponding to the inputs, given by $f = \G(x)$. The classifier $\C$ then outputs a softmax prediction distribution over the classes, and it is trained using a cross entropy (CE) loss on the labeled source data given by
\begin{equation}
    \LL_{sup} = \mathbb{E}_{(x,y) \sim \mathcal{D}^s} [- \log [\C(\G(x))]_{y}],
\end{equation}
\noindent where $y$ is the ground truth label corresponding to the source input $x$ and the expectation is taken over all the source data $\mathcal{D}^s$. However, since $P_s \neq P_t$, the classifier trained on source data does not transfer well to target samples, and an adversarial learning strategy~\cite{DANN, tzeng2017adversarial} is used to alleviate this issue. A domain discriminator $\D$ is trained using $\LL_{D}$ to classify between source and target, while $\G$ is simultaneously trained using $\LL_{adv}$ to generate features that confuse the discriminator:
\begin{eqnarray}
    \LL_{adv} &=& \mathbb{E}_{x \sim \mathcal{D}^t} [-\log \D(\G(x))], \\
    \LL_D &=& - \mathbb{E}_{x \sim \mathcal{D}^s} [\log \D(\G(x))]  \nonumber \\
    && - \mathbb{E}_{x \sim \mathcal{D}^t } [\log ( 1-\D(\G(x)))].
\end{eqnarray}

Min-max training between $\LL_D$ and $\LL_{adv}$ then yields domain invariant features. However, this is not enough to guarantee class specific alignment between source and target, so we present our proposed affinity matrix based adaptation next.

\subsection{Multi-Sample Contrastive (MSC) Loss}
\label{subsec:msc_loss}

%%%%%%%%%%%%%%%%%%%%%%%%%%%%%%%%%%%%%%%%%%%%%%%%
%%%%%%%%%%%%%%%%%%%%%%%%%%%%%%%%%%%%%%%%%%%%%%%%
\begin{figure}[tbp]
  \centering
    \includegraphics[width=.5\textwidth]{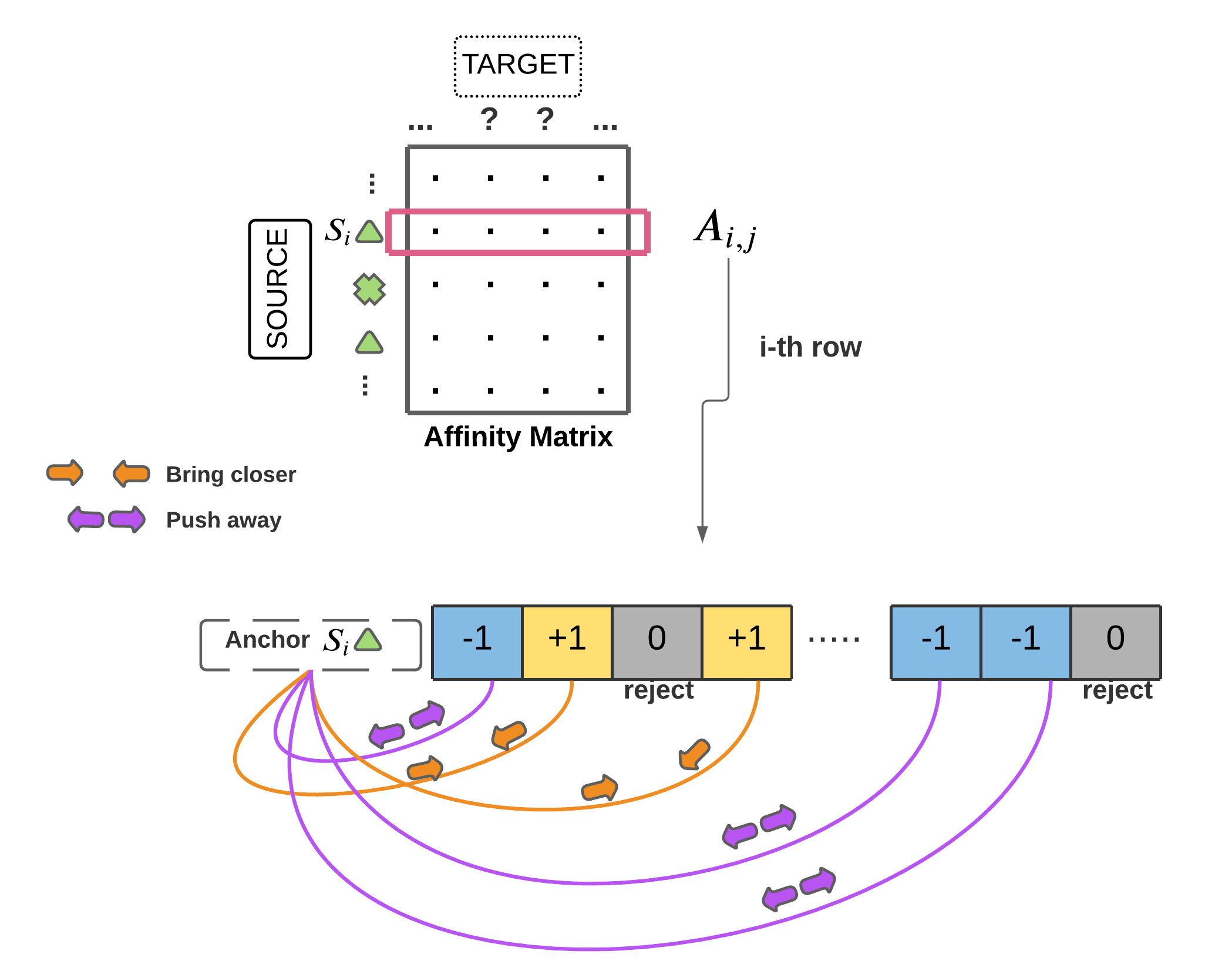}
  \caption{\bd{Illustration of our MSC loss.} $i^{th}$ row of affinity matrix $\Aff_i$ contains similarity information of $i^{th}$ source sample with every target sample. MSC loss uses these relations to attract positive sample from target while separating negative ones. }
  \label{fig:msc_loss_illustration}
    \vspace{-2pt}
\end{figure}
%%%%%%%%%%%%%%%%%%%%%%%%%%%%%%%%%%%%%%%%%%%%%%%%
%%%%%%%%%%%%%%%%%%%%%%%%%%%%%%%%%%%%%%%%%%%%%%%%

To enforce the class-level alignment constraint, we first find the sample level similarity scores among the source and target samples in a mini-batch and use them in our multi-sample contrastive (MSC) loss. However, we do not have labels for the target domain, so we follow a kNN based approach to assign each target sample in a mini-batch with a label belonging to nearby source samples. We then construct an affinity matrix $\Aff$, in which $\Aff_{ij}=1$ if the $i^{th}$ source sample from the mini-batch is similar to $j^{th}$ target sample from the mini-batch, $\Aff_{ij}=-1$ if it is not. This is explained in detail in \secref{subsec:affinity}. Assuming we have constructed such an affinity matrix $\Aff$, we use this information to construct positive and negative samples corresponding to a source sample $x_i$. Specifically, let $B_S$ and $B_T$ be the source and target batches respectively. Then, for each source sample $x_i \in B_s$, we identify the set of positive target pairs as $B_T^{i+} = \{x_j \in B_T | \Aff_{ij} = 1\}$, and negative pairs as $B_T^{i-} = \{x_j \in B_T | \Aff_{ij} = -1\}$. 
We then use this information to pull similar samples across source and target closer to each other, while pushing away dissimilar samples using our MSC loss given by:
\begin{IEEEeqnarray}{l}
    \LL_{MSC}^i = 
    \displaystyle -\log \frac{ \sum \limits_{j \in B_T^{i+}} e^{\phi(f_i , f_j)} }{ \sum \limits_{j \in B_T^{i+}} e^{\phi(f_i , f_j)} +  \sum \limits_{j \in B_T^{i-}} e^{\phi(f_i , f_j)} }, \IEEEeqnarraynumspace
    \label{eq:msc_loss}
\end{IEEEeqnarray} 
\noindent where $B_S$ and $B_T$ denote the source and target batches respectively, $f$ are the features computed as the output of $\G(x)$ and $\phi(.,.)$ is any metric that takes the features and outputs a similarity score. The overall loss is computed as the average across all the source samples from the mini-batch $B_S$:
\begin{equation}
   \LL_{MSC} = \frac{1}{|B_S|}  \sum_{i \in B_S} \LL_{MSC}^i.
\end{equation}
Empirically, we observe best results when using normalized inverse Euclidean distance~\cite{maaten2008visualizing} as the similarity metric $\phi$:
\begin{equation}
    \phi(f_i , f_j) = \frac{1}{1 + ||f_i - f_j ||^2}.
\end{equation}

\noindent This process is illustrated in \figref{fig:msc_loss_illustration}. Similar kind of contrastive loss is used for learning representations from unlabeled image and video in \cite{oord2018representation, gordon2020watching, he2020momentum, chen2020simple} where positives come from transformed versions of inputs unlike \Ours{}. {Furthermore, contrastive loss is shown to work well for large intra-class variations empirically in \cite{he2020momentum, InstDist} and theoretically in \cite{arora2019theoretical}. \Ours{} demonstrates similar benefits, while additionally accounting for possible domain gap between the positive and negative pairs}. From \eqref{eq:msc_loss}, we can observe that if $\Aff_{ij}=1$, indicating similar pairs, then the similarity metric needs to be higher to minimize the loss. Likewise, if $\Aff_{ij}=-1$, then the similarity score would be driven down to zero, as we require it to be. 
We now explain in detail the method to construct the affinity matrix $\Aff$. 

\begin{algorithm}
\captionsetup{labelfont={sc,bf}}
  \caption{Instance Affinity Based Adaptation during each iteration.}
  \label{alg:mscLoss}
\begin{algorithmic}[1]
\Require Class balanced mini-batches for source $B_S {\in} \mathcal{D}^s$ and randomly sampled target mini-batches $B_T {\in} \mathcal{D}^t$ 
\Require Feature extractor $\G(.)$ 
\Require Similarity metric $\phi(.,.)$
\State {$\Aff_{ij}=0$ $\forall i \in \{1,2..,|B_S|\}$, $j \in \{1,2..,|B_T|\}$} 
\For{$x_j$ in $B_T$} \Comment Construct affinity matrix
    \State $\hat{y}_j{=kNN(B_S, x_j)}$  (\secref{subsec:affinity})
    \For{$x_i$ in $B_S$}
        \State $\Aff_{ij}=1$ \textbf{if} $y_i = \hat{y}_j$ \textbf{else} $\Aff_{ij}=-1$
    \EndFor
\EndFor
\For{$x_j$ in $B_T$}
    \State {$\Gamma_{j}(x_j)$ (Eq~\eqref{eq:gamma}) \Comment{Compute Similarity Ratio}}
\EndFor
\State {$B_T^{F}=$ $Filter(B_T, \Gamma, \mu)$ \Comment{Select confident pseudo-labels using similarity ratio test.}
\State $Loss = MSC(B_S, B_T^{F}$) (Eq \eqref{eq:msc_loss}) \Comment{Compute MSC loss.} }
\end{algorithmic}
\end{algorithm}

\subsection{Constructing the Affinity Matrix}
\label{subsec:affinity}

Recall that the target dataset is completely unlabeled, so obtaining similarity scores is not trivial. Using source classifier to assign pseudo labels is an option, but it would be noisy during initial stages of training and empirically suboptimal (\secref{subsec:ablations}, \tabref{table:ablation_pseudolabel}). Instead, we rely on a k-nearest neighbor approach followed by a ratio test to assign confident target labels. For every target sample $x_j \in B_T$, we take the $k$ nearest neighbors ranked using the same similarity metric $\phi(f_j , .)$ from the source mini-batch. Then, the target sample is assigned to the class that is most common among its source neighbors and we populate the $j^{th}$ column of the affinity matrix $\Aff$ using this assignment. That is,
$$
\Aff_{ij}=\begin{cases}
			1, & \text{if } y_i=\hat{y}_j \\
            -1, & \text{if } y_i \neq \hat{y}_j.
		 \end{cases}
$$

\noindent Although the similarity and dissimilarity relations can now be directly read off the affinity matrix $\Aff$, we did not yet account for the fact that some of the psuedo labels can be noisy. 
After constructing the affinity matrix $\Aff$, we filter out possible noisy pseudo-labels. 
For this, we use a rejection based confidence measure commonly used in kNN literature based on a neighborhood similarity ratio test~\cite{delany2005generating, dasarathy1995nearest}. 
Denote using $N^l_j$ the set of like samples from source in the neighborhood of a target sample $x_j \in B_T$, given by $N^l_j : \{x_i \in B_S | y_i = \hat{y}_j \}$. Similarly, the set of unlike samples from source in the neighborhood is given by $N^u_j$, where $N^u_j : \{x_i \in B_S | y_i \neq \hat{y}_j \}$. We calculate the confidence score of a particular pseudo label prediction $\Gamma_j$ using the ratio of aggregate similarity between the sample and the like and unlike sets. That is,
% %
%%
\begin{equation}
    \Gamma_j = \frac{\sum_{x_i \in N^l_j} \phi(f_j , f_i) }{ \sum_{x_i \in N^u_j} \phi(f_j , f_i)  }.
    \label{eq:gamma}
\end{equation}

\noindent We then choose a sampling factor $\mu$, and select the top $\mu$ fraction of target samples and declare them to be confident, and for the rest of target samples, we put $A_{ij}=0$ and do not use them anymore in the MSC loss \eqref{eq:msc_loss}. For example, if sampling factor $\mu = .75$ with a batch size of 128, we select the top 96 target samples ranked based on their prediction confidence $\Gamma$. Since the number of unlike samples are generally much higher than the number of like samples, we only take the top $m$ samples in the summations in Eq~\eqref{eq:gamma} to balance the aggregate between the like and unlike sets. We chose m to be the maximum possible similar samples across datasets. In our case $m$ is equal to the size of each class in source mini-batch. This way, we will be left with pairwise similarity scores between pairs of source and target samples which pass the similarity ratio test. Further analysis of such a psuedo labeling procedure, including the sensitivity to the sampling factor $\mu$, is presented in \secref{subsec:ablations}. The complete algorithm is summarized in \AlgRef{alg:mscLoss}.
Although many prior works have considered a psuedo-labeling criterion for assigning target labels during training~\cite{chen2019progressive, xie2018learning}, the advantage we provide lies in the fact that our MSC loss takes sample level similarities with an explicit push-pull objective which is greatly useful to model finer category separation. Also, we have $O(n^2)$ psuedo-labels in each mini-batch of size $n$, so we will be left with a strong signal even after removing lesser confident predictions. 
{In contrast to \cite{liang2019exploring}, we extract kNN neighbors across source and target, calculate sample-sample as opposed to sample prototype relations for use in our MSC loss.}

Finally, when we randomly sample mini batches from source and target, it might so happen that some classes might not get picked in source, which is problematic. For example, some target samples might not have a corresponding true source sample leading to incorrect psuedo labels, or some source sample might get paired with a dissimilar target sample in our MSC loss in Eq~\eqref{eq:msc_loss}. To avoid this issue, we perform class balanced mini batch sampling only on the source dataset, in which we make sure that all classes have equal representation in all the sampled source mini batches $B_S$. Unlabeled target mini-batches are sampled randomly.

\section{Results and Analysis}
\label{sec:expmnts}

In this section, we conduct extensive experiments on multiple domain adaptation benchmarks to verify the effectiveness of ILA-DA approach. We next present the datasets used to evaluate our results, baselines methods we compared against, followed by results and discussion.

\begin{table}[!t]
    \renewcommand\arraystretch{0.8}
    \centering
    \small
    \resizebox{0.45\textwidth}{!}{
      \begin{tabular}{ccccc}
            \toprule
            Method & M $\xrightarrow{}$ U & U $\xrightarrow{}$ M & S $\xrightarrow{}$ M & Avg. \\
            \midrule
          Source Only & 76.7 & 63.4  & 67.1 & 69.1\\
          DANN~\cite{DANN} & 90.8 & 93.95 & 83.11 & 89.29\\
          ADDA~\cite{tzeng2017adversarial} & 89.4 & 90.1 & 76.0 & 85.2 \\
          DSN~\cite{bousmalis2016domain} & 91.3 & - & 82.7 &- \\
          ATT~\cite{saito2017asymmetric} & - & - & 85.0 & - \\
        ILA-DA (with DANN) & 92.43 & 97.32 & 91.84 & 93.83  \\
    	\midrule
      	CDAN~\cite{CDAN} & 93.9 & 96.9 & 88.5 & 93.1\\
        ILA-DA (with CDAN) & \textbf{94.87} & \textbf{97.47} & \textbf{92.30} & \textbf{94.88}\\
            \bottomrule
          \end{tabular}}
    %   }
      \caption{Accuracy (\%) on Digits for unsupervised domain adaptation. Results shown for a value of $k = 3$ and $\mu = 0.75$.}
      \label{table:digits}
    \vspace{-8pt}
\end{table}

\begin{table*}
  \centering
  \rowcolors{2}{gray!25}{white}

  \begin{tabular}{@{} l *{14}{c} @{}} 
    \toprule  \\[-1em]
    Method && A $\rightarrow$ W && D $\rightarrow$ W && W $\rightarrow$ D && A $\rightarrow$ D && D $\rightarrow$ A && W $\rightarrow$ A && Avg. \\
    \midrule
    ResNet-50 && 68.4 && 96.7 && 99.3 && 68.90 && 62.50 && 60.70 && 76.1 \\
    DAN~\cite{long2015learning} && 80.5 && 97.1 && 99.6 && 78.6 && 63.6 && 62.8 && 80.4 \\
    RTN~\cite{long2016unsupervised} && 84.5 && 96.8 && 99.4 && 77.5 && 66.2 && 64.8 && 81.6 \\
    DANN~\cite{DANN} && 82.0 && 96.9 && 99.1 && 79.7 && 68.2 && 67.4 && 82.2 \\
    ADDA~\cite{tzeng2017adversarial} && 86.2 && 96.2 && 98.4 && 77.8 && 69.5 && 68.9 && 82.9 \\
    MCD~\cite{saito2018maximum} && 88.6 && 98.5 && 100.0 && 92.2 && 69.5 && 69.7 && 86.5 \\
    SimNet~\cite{pinheiro2018unsupervised} && 88.6 && 98.2 && 99.7 && 85.3 && 73.4 && 71.8 && 86.2 \\
    GTA~\cite{sankaranarayanan2018generate} && 89.5 && 97.9 && 99.8 && 87.7 && 72.8 && 71.4 && 86.5 \\
    CDAN~\cite{CDAN} && 93.1 && 98.2 && \textbf{100.0} && 89.8 && 70.1 && 68.0 && 86.6 \\
    CDAN+E~\cite{CDAN} && 94.1 && 98.6 && \textbf{100.0} && 92.9 && 71.0 && 69.3 && 87.7 \\
    DAA~\cite{kang2018deep} && 86.8 && 99.3 && \textbf{100.0} && 88.8 && \textbf{74.3} && 73.9 && 87.2 \\
    SAFN~\cite{xu2019larger} && 88.8 && 98.4 && 99.8 && 87.7 && 69.8 && 69.7 && 85.7 \\
    MADA~\cite{pei2018multi} && 90.0 && 97.4 && 99.6 && 87.8 && 70.3 && 66.4 && 85.2 \\
    \midrule
    ILA-DA (with DANN) && 89.05 && 98.49 && \textbf{100.0} && 86.55 && 69.47 && 69.72 && 85.54 \\
    ILA-DA (with CDAN)  && \textbf{95.72} && \textbf{99.25} && \textbf{100.0} && \textbf{93.37} && 72.10 && \textbf{75.40} && \textbf{89.30}\\
    \bottomrule
  \end{tabular}

  \caption{\textbf{Office-31 dataset} Results for domain adaptation on Office-31 adaptation setting using Resnet-50 for 6 transfer tasks among three domains: Amazon (A), Webcam (W) and Dslr (D). Our method shows consistent improvements. All the baselines as well as ours use ResNet-50 as the backbone architecture. Results shown for $k = 5$ and $\mu=0.67$.}
  \label{tab:office31_results}
  \vspace{-12pt}
\end{table*}

\subsection{Experimental Details}

\paragraph{Datasets} We investigate the performance of our model on three different kinds of benchmark datasets used for domain adaptation, namely Digits, Office-31 and Birds-31.

\vspace{0.15cm}
\noindent{\em Digits}. We use SVHN, MNIST and USPS consisting of images of digits $0-9$.
We explore the adaptation tasks between \textbf{MNIST $\rightarrow$ USPS}, \textbf{USPS $\rightarrow$ MNIST} and \textbf{SVHN $\rightarrow$ MNIST}. 

\vspace{0.15cm}
\noindent{\em Office-31}. This setting consists of images from 31 categories 
from three different domains, 
namely Amazon (\textbf{A}), Webcam (\textbf{W}) and DSLR (\textbf{D}).
We show results for all the 6 task pairs \textbf{A → W, D → W, W → D, A → D, D → A} and \textbf{W → A}.  Following prior works, we report results on the complete unlabeled examples of the target domain. 

\vspace{0.15cm}
\noindent{\em Birds-31}. This dataset is recently proposed by \cite{PAN} for fine grained adaptation consisting of different types of birds. We use it to verify our argument that our MSC loss performs efficiently even with datasets that possess large intra-class and small inter-class variation. It consists of three domains, namely, 1848 images from  CUB-200-2011 (\textbf{C})~\cite{cub}, 2988 images from NABirds (\textbf{N})~\cite{NaBirds} and 2857 images from iNaturalist2017 (\textbf{I}) datasets from the 31 common classes among the three. We show the adaptation results on six transfer tasks formed from three domains: \textbf{C → I, I → C, I → N, N → I, C → N} and \textbf{N → C}. 

\vspace{0.15cm}
\noindent\bd{Training details}. Following prior works~\cite{CDAN, pinheiro2018unsupervised}, we use LeNet architecture for digits and use ResNet-50 (pretrained on Imagenet) as the feature extractor $\G$ for the Office-31 and Birds-31 datasets, while the classifier $\C$ is made up of fully connected layers. For achieving training stability, we observe that it is essential to pretrain the model on the labeled source dataset for a few iterations before introducing our constrastive loss. 

We use mini-batch SGD with a learning rate of 0.001 for Office and 0.03 for birds. For the classifier we multiply the learning rate by 10. We use a similar annealing strategy as used in \cite{DANN}. 
Further details on the hyperparameter settings are presented in the supplementary material. 

To illustrate the benefits of the proposed MSC loss, we employ it on top of two competing adaptation benchmarks in DANN~\cite{DANN} and CDAN~\cite{CDAN}, while noting that our loss is general and applicable in combination with any adversarial adaptation approach. For experiments with DANN, we replace the adversarial loss with a gradient reversal layer.

\vspace{0.1cm}
\noindent\bd{Baselines}. We focus our comparison against works which use adversarial learning strategy to perform global domain level alignment such as \textbf{DAN}~\cite{long2015learning}, \textbf{RTN}~\cite{long2016unsupervised}, \textbf{ADDA}~\cite{tzeng2017adversarial}, \textbf{GTA}~\cite{sankaranarayanan2018generate}, \textbf{DAA}~\cite{kang2018deep} and \textbf{CDAN}~\cite{CDAN} as well as works which perform class aware alignment such as \textbf{MCD}~\cite{saito2018maximum}, \textbf{SimNet}~\cite{pinheiro2018unsupervised}, \textbf{MADA}~\cite{pei2018multi}. For Birds-31, we additionally verify our result with prior fine grained adaptation work, \textbf{PAN}~\cite{PAN}. 
Finally, we have \textbf{ILA + DANN}, which is using \Ours{} approach on top of DANN and \textbf{ILA + CDAN} which uses \Ours{} in combination with CDAN.
We compare the task-wise accuracies and also report the average accuracies across all the transfer tasks. Our training and evaluation scripts are publicly released.

\begin{table*}
  \centering
  \resizebox{0.85\textwidth}{!}{
  \rowcolors{2}{gray!25}{white}
  \begin{tabular}{@{} l *{14}{c} @{}} 
    \toprule  \\[-1em]
    Method && C $\rightarrow$ I && I $\rightarrow$ C && I $\rightarrow$ N && N $\rightarrow$ I && C $\rightarrow$ N && N $\rightarrow$ C && Avg. \\
    \midrule
    ResNet-50 && 64.25 && 87.19 && 82.46 && 71.08 && 79.92 && 89.96 && 79.14 \\
    DAN~\cite{long2015learning} && 63.9 && 85.86 && 82.91 && 70.67 && 80.64 && 89.40 && 78.90 \\
    DANN~\cite{DANN} && 64.59 && 85.64 && 80.53 && 71.00 && 79.37 && 89.53 && 78.44 \\
    JAN~\cite{long2017deep} && 63.69 && 86.29 && 83.34 && 71.09 && 81.06 && 89.55 && 79.17 \\
    ADDA~\cite{tzeng2017adversarial} && 63.03 && 87.26 && 84.36 && 72.39 && 79.69 && 89.28 && 79.33 \\
    MADA~\cite{pei2018multi} && 62.03 && 89.99 && 87.05 && 70.99 && 81.36 && 92.09 && 80.50 \\
    MCD~\cite{saito2018maximum} && 66.43 && 88.02 && 85.57 && 73.06 && 82.37 && 90.99 && 81.07 \\
    CDAN~\cite{CDAN} && 68.67 && 89.74 && 86.17 && 73.80 && 83.18 && 91.56 && 82.18 \\
    SAFN~\cite{xu2019larger} && 65.23 && 90.18 && 84.71 && 73.00 && 81.65 && 91.47 && 81.08 \\
    PAN~\cite{PAN} && 69.79 && 90.46 && 88.10 && 75.03 && 84.19 && 92.51 && 83.34 \\
    \midrule
    ILA-DA (with DANN) && 69.55 && 93.13 && 87.15&& 74.69 && 83.40&& 93.89&& 83.63 \\
    ILA-DA (with CDAN)  && \textbf{72.77} && \textbf{93.83} && \textbf{90.36} && \textbf{78.09} && \textbf{86.58} && \textbf{94.53} && \textbf{86.03}\\
    \bottomrule
  \end{tabular}
  }
  \caption{Results for domain adaptation on fine-frained adaptation setting, shown for 3 challenging datasets, namely CUB-200-2011 (C), iNaturalist2017 (I)  and NABirds (N). We perform consistently better than all other methods by explicitly modeling the finegrained nature of the adaptation process. All the methods use ResNet-50 pretrained on ImageNet. All the baseline numbers taken from \cite{PAN}. Results shown for $k = 3$ and $\mu=0.33$.}
  \label{tab:birds31_result}
  \vspace{-12pt}
\end{table*}

\subsection{Comparison with State-of-the-art}

\paragraph{Digits} In \tabref{table:digits}, we show the results for adaptation using our method with MSC loss. We observe that we outperform prior methods by a significant margin when we use CDAN in combination with \Ours{}. On \textbf{MNIST → USPS} we observe an improvement from $90.8$ to $92.43$ while using \Ours{} {+} DANN and $93.9$ to $94.87$ with \Ours{} {+} CDAN, indicating the usefulness of our MSC loss for improving existing methods for domain adaptation.  Similar improvements can be observed for all other dataset settings as well, for instance, accuracy goes up from $88.5$ to $92.30$ in the case of \textbf{SVHN → MNIST} using \Ours{} with CDAN. 

\vspace{0.15cm}
\noindent \bd{Office-31} We present results on the 6 transfer tasks on Office-31, including their average, in \tabref{tab:office31_results}. We observe that we achieve an accuracy of $89.30\%$ on the average, outperforming all the competing baselines, which includes prior works that perform global domain alignment~\cite{long2016unsupervised, tzeng2017adversarial}, as well as those that model finer class separation~\cite{saito2018maximum, pinheiro2018unsupervised, pei2018multi} like us, highlighting the advantages of our MSC loss in comparison to competing approaches.
Finally, to testify that our loss is generally applicable, we show that it improves accuracy over both the approaches DANN~\cite{DANN} and CDAN~\cite{CDAN}, consistently over all the tasks (by $3.3\% $ and $2.7\%$ on average, respectively). This result underlines the necessity for our sample aware class-specific transfer in addition to global domain alignment. 

\vspace{0.15cm}
\noindent \bd{Birds-31} The difficulty in this setting lies in the fact that birds from same class but different domains look quite distinct, sometimes more different than images from another class. We verify the results on all 6 transfer tasks on Birds-31 dataset in \tabref{tab:birds31_result}, and show that \Ours{} outperforms prior works across all the tasks. Due to the intra-class variation in the dataset and small inter-class distances, prior works that rely on global alignment objectives~\cite{long2015learning, long2017deep, tzeng2017adversarial} do not perform any better than a source-only model (\textit{ResNet-50} baseline), possibly because they suffer from negative alignment. However, our MSC loss explicitly accounts for the instance level relations to model category separation, which pulls similar samples from both datasets closer while pushing away dissimilar ones. As a result, we improve the accuracy over DANN on all the tasks, and average accuracy from $78.44\%$ to $83.63\%$. In fact, with an average accuracy of $86.03\%$  we achieve the \textit{new state-of-the-art result} using \Ours{} in combination with CDAN. More remarkably, \Ours{}+CDAN even outperform PAN~\cite{PAN}, that is specifically designed for fine-grained adaptation by roughly $3\%$ without demanding access to any label structure and class hierarchy during training unlike \cite{PAN}, which highlights the usefulness of modeling instance level loss for challenging adaptation problems.

\subsection{Ablations and Analysis}
\label{subsec:ablations}

\noindent \bd{Importance of MSC Loss} We testify the effectiveness of the proposed multisample contrastive loss in modeling the instance level relations by comparing it to another commonly used metric loss, namely triplet loss. We replace the loss used in Eq~\eqref{eq:msc_loss} by triplet loss, by deriving positives and negatives from the affinity matrix. We use similarity metric $\phi(.)$, and choose the nearest negative sample and farthest positive sample as hard negative and hard positive respectively. From \tabref{table:triplet_loss}, we first observe that both triplet loss as well as MSC loss improve over {CDAN} baseline, which indicates the usefulness of adding metric learning losses over {adversarial methods for better alignment. Further, we also observe that replacing MSC loss by triplet loss leads to drop in accuracy from $93.37\%$ to $90.20\%$ on A$\rightarrow$D and from $75.40\%$ to $73.94\%$ on W$\rightarrow$A settings on Office-31 dataset.} From this, we conclude that for improving domain adaptation, modeling multiple instance relations at once using MSC loss is simpler and more powerful than triplet loss. 

\vspace{0.15cm}
\noindent \bd{Choice of psuedo-labeling} In proposed \Ours{}, the psuedo labeling process for the target examples is driven by finding the $k$ nearest source neighbors in the feature space. Alternatively, we can directly use the source classifier predictions as psuedo labels~\cite{xie2018learning, pei2018multi}. To tease out the differences between these alternatives, we compare against such a classifier based psuedo labeling method {which filters the target samples using softmax scores as an indicator for the prediction confidence, in \tabref{table:ablation_pseudolabel} .} We observe that our kNN based approach provides significant benefit over the classifier based counterpart on all the tasks, with a {$2.62\%$} boost in accuracy on average.

% Tables for analysis

\begin{table*}
\centering
\hfill
\begin{subtable}[t]{.23\textwidth}
\resizebox{\textwidth}{!}{
    \begin{tabular}{cccccc}
          \toprule
          
            Method & A$\xrightarrow{}$D & W$\xrightarrow{}$A & Avg. \\
            \midrule
           
           CDAN\cite{CDAN}  & 89.8 & 68.0 & 86.6 \\
            \midrule

            Triplet + CDAN & 90.20  & 73.94 & 87.63\\
            MSC + CDAN &  \textbf{93.37} & \textbf{75.40} & \textbf{89.30}\\
            
            \bottomrule
           
    \end{tabular}}
\captionsetup{width=.9\textwidth}
\caption{Comparison of triplet Loss vs. MSC Loss for metric learning. Results shown for A$\xrightarrow{}$D and W$\xrightarrow{}$A tasks and avg. of all 6 tasks from Office-31 dataset.}
\label{table:triplet_loss}
\end{subtable}
\hfill
\begin{subtable}[t]{.23\textwidth}
\resizebox{\textwidth}{!}{
\begin{tabular}{cccccc}
        \toprule
        Method & A$\xrightarrow{}$D & W$\xrightarrow{}$A & Avg. \\
        \midrule
        CDAN\cite{CDAN} & 89.8 & 68.0 & 86.6  \\
        \midrule
        classifier based & 88.35 & 70.11 &  86.68 \\
       
    	kNN based  & \textbf{93.37}& \textbf{75.40} & \textbf{89.30}\\
  
        \bottomrule
    \end{tabular}
    }
\captionsetup{width=.9\textwidth}
\caption{ Comparison of kNN vs. Classifier based psuedo-labeling schemes. Results shown for A$\xrightarrow{}$D and W$\xrightarrow{}$A tasks and avg. of all 6 tasks from Office-31 dataset for $k=5$, $\mu$=0.67.}
\label{table:ablation_pseudolabel}
\end{subtable}
\hfill
\begin{subtable}[t]{.23\textwidth}
\resizebox{\textwidth}{!}{
\begin{tabular}{cccc}
        \toprule

        Method & A$\xrightarrow{}$D & W$\xrightarrow{}$A & Avg.  \\
        \midrule
        CDAN\cite{CDAN}  & 89.8 & 68.0 & 86.6 \\
        \midrule
         \Ours{} , $\mu{=}0.33$  &  90.75   &  71.95 & 87.91       \\
          \Ours{} , $\mu{=}0.50$ &    91.95  &  74.33 & 88.53     \\
           \Ours{} , $\mu{=}0.67$ &   \textbf{93.37}   &  \textbf{75.40} &       \textbf{89.30} \\
           \Ours{},  $\mu{=}1.00$ &  92.33  & 70.39&      87.92 \\

        \bottomrule
    \end{tabular}}
\captionsetup{width=.9\textwidth}
\caption{Effect of sampling fraction $\mu$. Results shown for A$\xrightarrow{}$D and W$\xrightarrow{}$A tasks and avg. of all 6 tasks from Office-31 dataset for $k = 5$.}
\label{table:ablation_mu}
\end{subtable}
\hfill
\begin{subtable}[t]{.23\textwidth}
\resizebox{\textwidth}{!}{
\begin{tabular}{cccc}
        \toprule
        Method & A$\xrightarrow{}$D & W$\xrightarrow{}$A & Avg.  \\
        \midrule
        CDAN\cite{CDAN} & 89.8  & 68.0 & 86.6 \\
        \midrule
         \Ours{}, $k{=}1$ &   91.96   &  69.93   &  87.46     \\
          \Ours{}, $k{=}3$ &  91.16    & 75.15  &   88.87   \\
           \Ours{}, $k{=}5$ & \textbf{93.37}    &  \textbf{75.40}  & \textbf{89.30  }     \\
       
        \bottomrule
    \end{tabular}}
\captionsetup{width=.9\textwidth}
\caption{Effect of number of neighbors $k$ used in psuedo-labeling. Results are shown for A$\xrightarrow{}$D and W$\xrightarrow{}$A tasks and avg. of all 6 tasks from Office-31 dataset.}
\label{table:ablation_K}
\end{subtable}
\vspace{-4pt}
\captionsetup{width=\textwidth}
\caption{Ablations into the proposed \Ours{}. \tabref{table:triplet_loss} shows comparison between triplet loss and proposed MSC loss. \tabref{table:ablation_pseudolabel} shows comparison between kNN and classifier based psuedo labeling schemes. \tabref{table:ablation_mu} and \tabref{table:ablation_K} summarize the effect of $\mu$ and $k$ respectively.}
\vspace{-4pt}
\label{table:ablations}
\end{table*}

\vspace{0.15cm}
\noindent \bd{Effect of sampling factor} We investigate the effect of the sampling parameter $\mu$, used to threshold the similarity ratio $\Gamma$ in Eq~\eqref{eq:gamma}. Intuitively, a very high value of $\mu$ would lead to many noisy psuedo labels being accepted leading to poor optimization, while a low value would eliminate even moderately confident positives which could be useful training signal. In fact, from \tabref{table:ablation_mu} we observe that  a value of $\mu=0.67$ is optimal, which corresponds to accepting the top two-thirds of the psuedo-label predictions. 

\vspace{0.15cm}
\noindent \bd{Effect of $k$} We show the effect of $k$ in the kNN process in \tabref{table:ablation_K}. We observe that the average accuracy on Office-31 dataset is highest for $k=5$. We provide further analysis on the influence of $k$ in supplementary material. In general, we find that a value of $k > 1$ is beneficial for reliable psuedo-labeling, as it helps handle noisy predictions around classifier boundaries.

%% Qualitative Figures
\begin{figure}
    \begin{center}
    \begin{subfigure}[b]{0.15\textwidth}
        \centering
        \includegraphics[width=\textwidth]{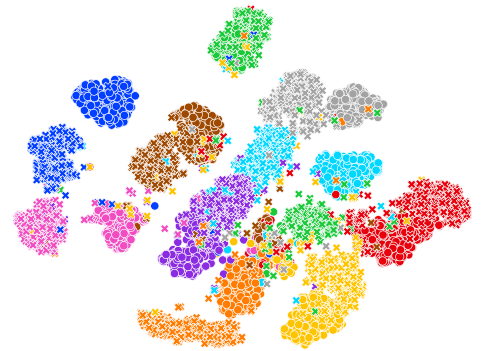}
    \end{subfigure}
    \hfill
    \begin{subfigure}[b]{0.15\textwidth}
        \centering
        \includegraphics[width=\textwidth]{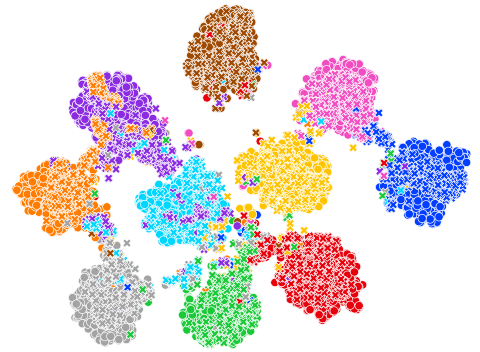}
    \end{subfigure}
    \hfill
    \begin{subfigure}[b]{0.15\textwidth}
        \centering
        \includegraphics[width=\textwidth]{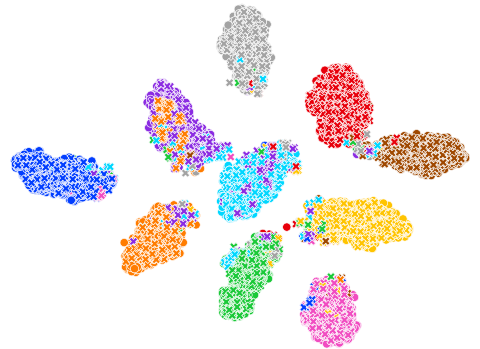}
    \end{subfigure}
    
    %%%%%
    %%%%%
    
    \begin{subfigure}[b]{0.15\textwidth}
        \centering
        \includegraphics[width=\textwidth]{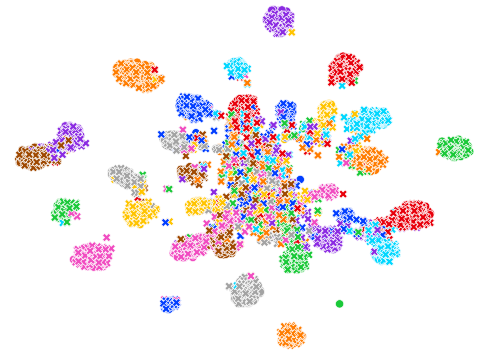}
        \caption{No Adapt}
        \label{fig:tsne_noadapt}
    \end{subfigure}
    \hfill
    \begin{subfigure}[b]{0.15\textwidth}
        \centering
        \includegraphics[width=\textwidth]{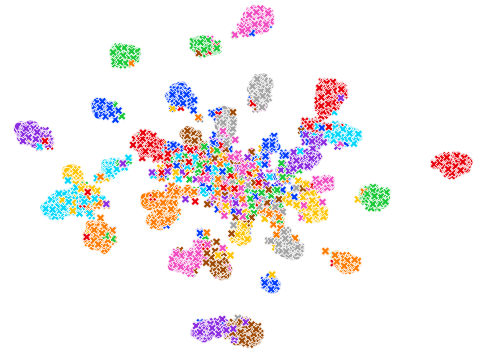}
        \caption{DANN}
        \label{fig:tsne_dann}
    \end{subfigure}
    \hfill
    \begin{subfigure}[b]{0.15\textwidth}
        \centering
        \includegraphics[width=\textwidth]{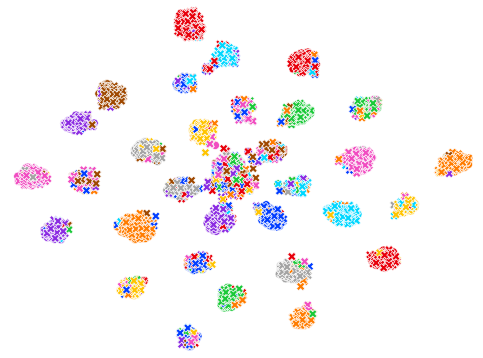}
        \caption{\Ours{}}
        \label{fig:tsne_ila}
    \end{subfigure}
    %%%%%%%%%%%
    %%%%%%%%%%%
    \end{center}
    \vspace{-16pt}
    \caption{tSNE visualizations of source and target features belonging to 10 classes from Digits (S$\xrightarrow{}$M ) (top row) and 31 classes from birds (N$\xrightarrow{}$I) (bottom row). Here, \ref{fig:tsne_noadapt} shows tSNE with no adaptation. While DANN~\cite{DANN} (\ref{fig:tsne_dann}) is only successful in domain alignment, our proposed \Ours{} approach additionally improves category separation on target domain (\ref{fig:tsne_ila}).
    }
    \label{fig:tSNEVis}
    \vspace{-4pt}
\end{figure}

\begin{figure}[tbp]
  \centering
    \includegraphics[width=.48\textwidth]{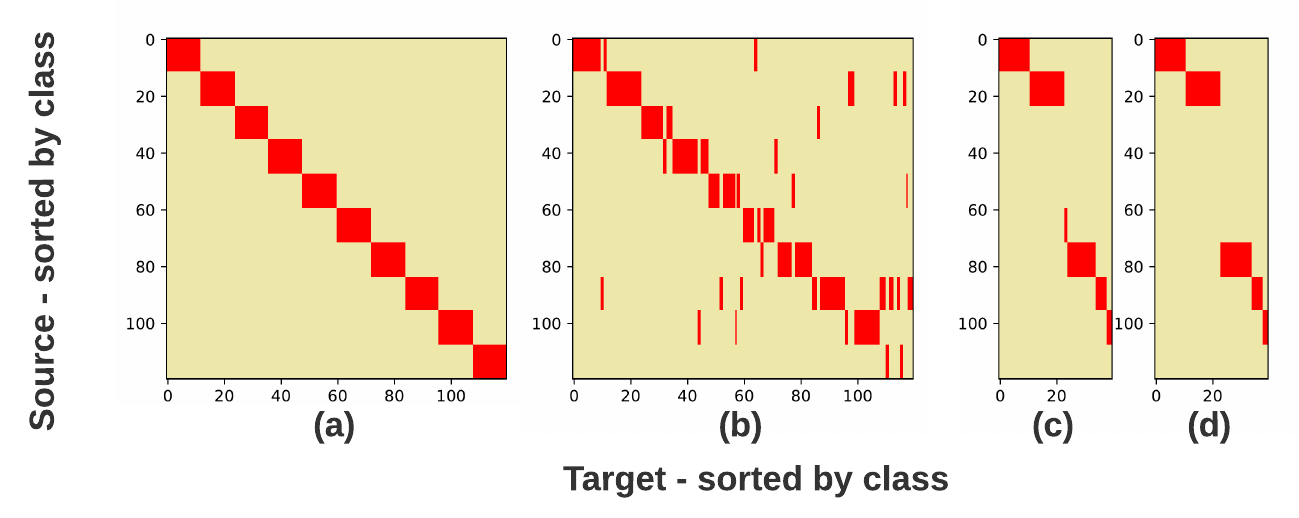}
    \vspace{-4pt}
    \caption{Visualization of Affinity matrix $\Aff$ consisting of similarity relations between the source and target for a subset of samples. The computed affinity matrix (b) is close to the ground truth affinity matrix (a), and we further close the gap by efficiently filtering wrong predictions (ground truth in (c) and (d)). Results shown for task M$\xrightarrow{}$U from Digits at 40-th epoch during training. }
  \label{fig:ablations_am_viz}
  \vspace{-4pt}
\end{figure}

\vspace{0.15cm}
\noindent \bd{Visualizing the Affinity Matrix} We visualize the affinity matrix $\Aff$ in \secref{subsec:affinity} to get an idea of the reliability of predicted pseudo-labels. For a mini-batch of size $120$, we plot the $120 {\times} 120$ affinity matrix $\Aff$ in \figref{fig:ablations_am_viz}, grouped by the class ordering. Here, (a) is the affinity matrix constructed using the ground truth similarities. We observe that the unfiltered affinity matrix in (b) already does a good job in accurately predicting the similarity (red , +1) and dissimilarity(yellow , -1) relations between source and target. Furthermore, we filter out noisy pseudo labels using our filtering approach discussed in \secref{subsec:affinity}, and find that the affinity matrix after filtering (shown in (d)) is much more closer to ground truth affinity matrix, in (c), which verifies the robustness of our pseudo labeling approach. 

\vspace{0.15cm}
\noindent \bd{Feature visualization} We provide the tSNE visualization of the learned features for two different dataset settings in \figref{fig:tSNEVis}. On both these settings, we observe better domain alignment as well as target category separation using \Ours{}. Note that although DANN does succeed in aligning the source and target domains, it does not necessarily produce discriminative features, which is addressed by \Ours{}.

\vspace{-4pt}
\section{Conclusion}

In this work, we leverage principles from metric learning to improve domain adaptation. 
We propose an affinity matrix based approach, \Ours{}, that uses a multi sample contrastive loss to explicitly model instance level interactions across source and target. We show that this helps in improving category separation while preventing negative alignment.
The proposed approach is general, and can be easily applied on top of any existing adversarial adaptation method. We show numerical results on various challenging benchmark datasets and perform favorably against many existing adaptation methods.

\vspace{0.1cm}
\noindent\bd{Limitations and Future Work}. As with any method that extracts pairwise similarities, the process of constructing the affinity matrix at each iteration is memory intensive. Given current limits on memory, our model may handle a reasonable number of categories across source and target. In future work, we aim to devise newer sampling strategies for affinity matrix construction that allow handling much larger number of classes.

\vspace{0.1cm}
\small
\noindent\bd{Acknowledgments}
This work was supported by NSF CAREER 1751365, an IPE Data Analytic Governance and Accountability Fellowship and NSF CHASE-CI.
\normalsize

{\small
\bibliographystyle{ieee_fullname}
% \bibliography{ref}
% \bibliography{egbib}
% \bibliography{references}
}

\end{document}